\documentclass[a4paper,twoside]{article}

\usepackage{epsfig}
\usepackage{subcaption}
\usepackage{calc}
\usepackage{amssymb}
\usepackage{amstext}
\usepackage{amsmath}
\usepackage{amsthm}
\usepackage{multicol}
\usepackage{pslatex}
\usepackage{apalike}
\usepackage{algorithm2e}
\usepackage[bottom]{footmisc}
\usepackage{csquotes}
\let\bibhang\relax
\usepackage{natbib}
\usepackage{SCITEPRESS}% Please add other packages that you may need BEFORE the SCITEPRESS.sty package.

\begin{document}

\title{Non-invasive Growth Monitoring of Small Freshwater Fish in Home Aquariums via Stereo Vision}

\author{\authorname{Clemens Seibold\sup{1,2}\orcidAuthor{0000-0002-9318-5934}, 
Anna Hilsmann\sup{2}\orcidAuthor{0000-0002-2086-0951} and 
Peter Eisert\sup{1,2}\orcidAuthor{0000-0001-8378-4805}}
\affiliation{\sup{1}Humboldt University of Berlin, Berlin, Germany}
\affiliation{\sup{2}Fraunhofer HHI, Berlin, Germany}
\email{\{clemens.seibold, anna.hilsmann, peter.eisert\}@hhi.fraunhofer.de}
}


\abstract{Monitoring fish growth behavior provides relevant information about fish health in aquaculture and home aquariums. Yet, monitoring fish sizes poses different challenges, as fish are small and subject to strong refractive distortions in aquarium environments. Image-based measurement offers a practical, non-invasive alternative that allows frequent monitoring without disturbing the fish. In this paper, we propose a non-invasive refraction-aware stereo vision method to estimate fish length in aquariums. Our approach uses a YOLOv11-Pose network to detect fish and predict anatomical keypoints on the fish in each stereo image. A refraction-aware epipolar constraint accounting for the air-glass-water interfaces enables robust matching, and unreliable detections are removed using a learned quality score. A subsequent refraction-aware 3D triangulation recovers 3D keypoints, from which fish length is measured. We validate our approach on a new stereo dataset of endangered Sulawesi ricefish captured under aquarium-like conditions and demonstrate that filtering low-quality detections is essential for accurate length estimation. The proposed system offers a simple and practical solution for non-invasive growth monitoring and can be easily applied in home aquariums.}

\onecolumn \maketitle \normalsize \setcounter{footnote}{0} \vfill

\section{\uppercase{Introduction}}
\label{sec:introduction}

Monitoring fish growth provides important information about the welfare and development of fish. It is a valuable tool in the fishing industry~\citep{Meng25FishSurvey}, as well as in home aquaculture and in organizations dedicated to the cultivation of fish to preserve marine biodiversity~\citep{1000arten}. 
Anomalies in the growth pattern can indicate stress, diseases, improper nutrition, or genetic mutations. Early detection of anomalies through growth monitoring can enable timely interventions to address these issues.

However, routine measurements are difficult to perform, because manual handling can harm the fish and, like other regular health inspections, is very time consuming. 
A non-invasive fish measurement approach using a stereo-camera setup can therefore provide a practical alternative and can provide substantial benefits for biodiversity-focused fish farms and home aquarium owners. 

Underwater and aquarium scenarios are unique challenges for computer vision systems due to refraction and variable lighting. Common concepts, such as the epipolar constraint in stereo vision, break down due to the air-glass-water interface, meaning that most stereo computer vision libraries cannot be used without extensive adaptations. 
Additional challenges arise from the nature of aquarium fish, which are often small, partially transparent, fast-moving, and often hiding in aquarium planting or decor. 

We propose a simple and non-invasive stereo system that explicitly accounts for refraction at the air–glass–water interface to enable accurate fish size estimation in aquariums. Our approach leverages epipolar curves~\citep{Gedge2011} for robust correspondence search and uses a modified YOLOv11 architecture to predict bounding boxes and anatomical reference points for each fish. An additional YOLO head identifies detections unsuitable for size estimation. After filtering, matched detections are triangulated using a refraction-aware 3D reconstruction, and fish size is computed from the resulting 3D keypoints.

Our contributions are:
\begin{itemize}
\item We propose a non-invasive stereo-camera system for tracking fish growth in fish aquarium settings. 
\item We provide a novel annotated stereo-image dataset for fish detection and 3D localization.
\item We evaluate the precision of our proposed system on a new and challenging dataset of small freshwater fish.
\end{itemize}

In the following, we summarize related and used concepts in Section~\ref{sec:relatedWork}, followed by information about our dataset in Section~\ref{sec:dataset}. Our proposed system is presented in Section~\ref{sec:methodology}. In Section~\ref{sec:results} we evaluate our proposed systems and end our paper with a conclusion and future plans with Section~\ref{sec:conclusion}. 

\section{\uppercase{Related Work}}
\label{sec:relatedWork}
Refractions at the air-glass and glass-water interfaces violate the assumptions of the pinhole camera model. A physically accurate representation for this scenario is provided by the axial camera model, in which all rays forming the image intersect along a line rather than at a single point~\citep{Agrawal12}. Given the camera pose, the distances to the refractive surfaces, and the refraction indices, the exact point on this axis and its corresponding ray direction can be computed for every pixel, as described in~\cite{Luczynsski2017_pinax}. Using this 3D point on the axis and the refracted ray direction, triangulation can be performed analogous to conventional stereo systems. 
However, projecting a 3D point onto the 2D image plane requires solving a 12th degree equation when two refracting interfaces with different refractive indices are involved~\citep{Agrawal12}.
In traditional stereo setup with known camera intrinsics and extrinsics, the correspondence search for a point in one image can be restricted to the epipolar line in the other image~\citep{hartleyZisserman2003multiple}. Under refraction, these linear constraints no longer hold, and the search region must be defined by curves rather than lines. \cite{Gedge2011} proposed to describe epipolar curves using a piecewise linear approximation. These epipolar curves can be precomputed for a given existing stereo configuration based on the intrinsic and extrinsic parameters, including distance and angle to the refractive interface.

Accurate size measurement of animals or objects requires accurate detection and localization. For multi-object localization, the YOLO architecture has proven to be a powerful and light-weight solution~\citep{redmon2016YOLOorig}. Since its initial publication, numerous improvements and extensions have been introduced to increase robustness, accuracy, and performance~\citep{quan2025lightweightmultiframeintegrationrobust}. While the original architecture predicts bounding boxes only, additional heads have been proposed to estimate further object attributes such as segmentation masks or keypoints~\citep{yolo11_ultralytics}. 

Underwater fish length estimation using stereo vision has been studied in various contexts. \cite{Soom2025} proposed a method employing bounding boxes detected by a YOLO network for real-time fish size estimation in rivers. However, their approach does not consider depth variations, leading to inaccurate fish length estimations when fish are not parallel to the camera plane. \cite{Zhou2023} developed a segmentation-based length estimation technique that requires placing the fish in a controlled white-box environment, which is time-consuming and causes additional stress. Their method employs GrabCut with a manually selected seed point and derives length from oriented bounding boxes around the segmentation. \cite{Coronel2024} proposed a single-camera method that depends on a controlled setup, placing individual fish shallow-water box with a fixed distance to the camera. The fish is segmented using adaptive thresholding and the segmentation’s length and the resulting shape is used to estimate biomass with a multilayer perceptron. \cite{Ubina2022} developed a length measurement based on a Mask R-CNN~\citep{He2017} for segmentation followed by a stereo matching of the detected fish, enabling fish length monitoring in natural environment.  

\begin{figure}[!htb]
    \centering
    \includegraphics[width=0.95\linewidth]{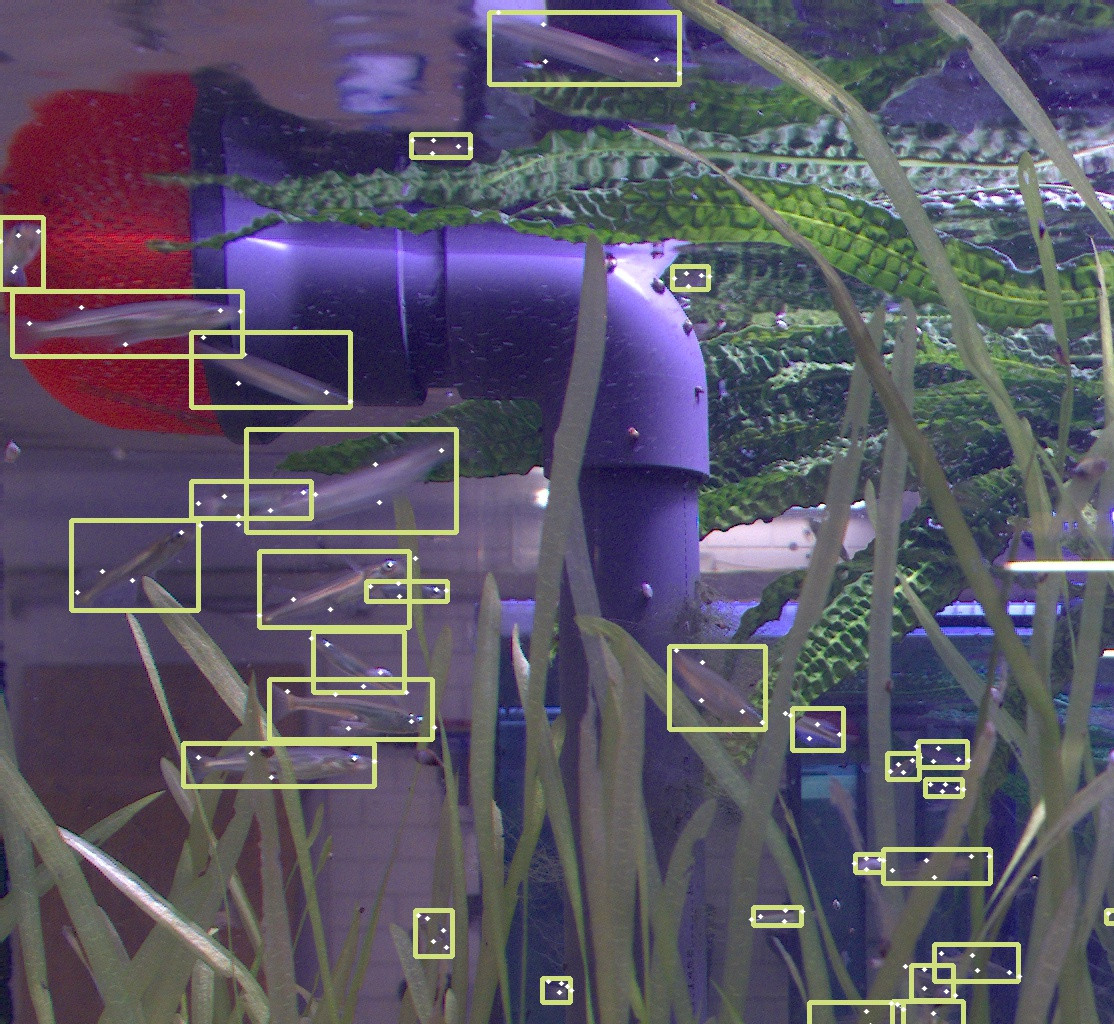}
    \caption{Example from our dataset. This image shows the upper left quadrant of a frame from our dataset. The fish only cover a small area of the image and partly hide between plants.}
    \label{fig:exFrame}
\end{figure}

In contrast, our approach targets small freshwater fish in crowded aquariums. To overcome challenges posed by fish size, density, and variable illumination, we propose a learned quality assessment and additional filtering strategies to discard poorly estimated stereo samples and obtain length measurements. Figure~\ref{fig:exFrame} shows the upper left quadrant of an image from our dataset, illustrating the difficulty of localizing and measuring small freshwater fish in aquariums.

\section{\uppercase{Dataset}}
\label{sec:dataset}
We recorded a fish tank containing the endangered Sulawesi ricefish with a Stereo setup with two Basler a2A2448-75ucPRO cameras equipped with 6 mm lenses,. This species grows up to only approximately 80 mm~\citep{Kottelat1993} and has a partially transparent caudal fin, which makes detection and measurement challenging. 

\begin{figure*}[!htb]
    \centering
    \includegraphics[width=0.475\linewidth]{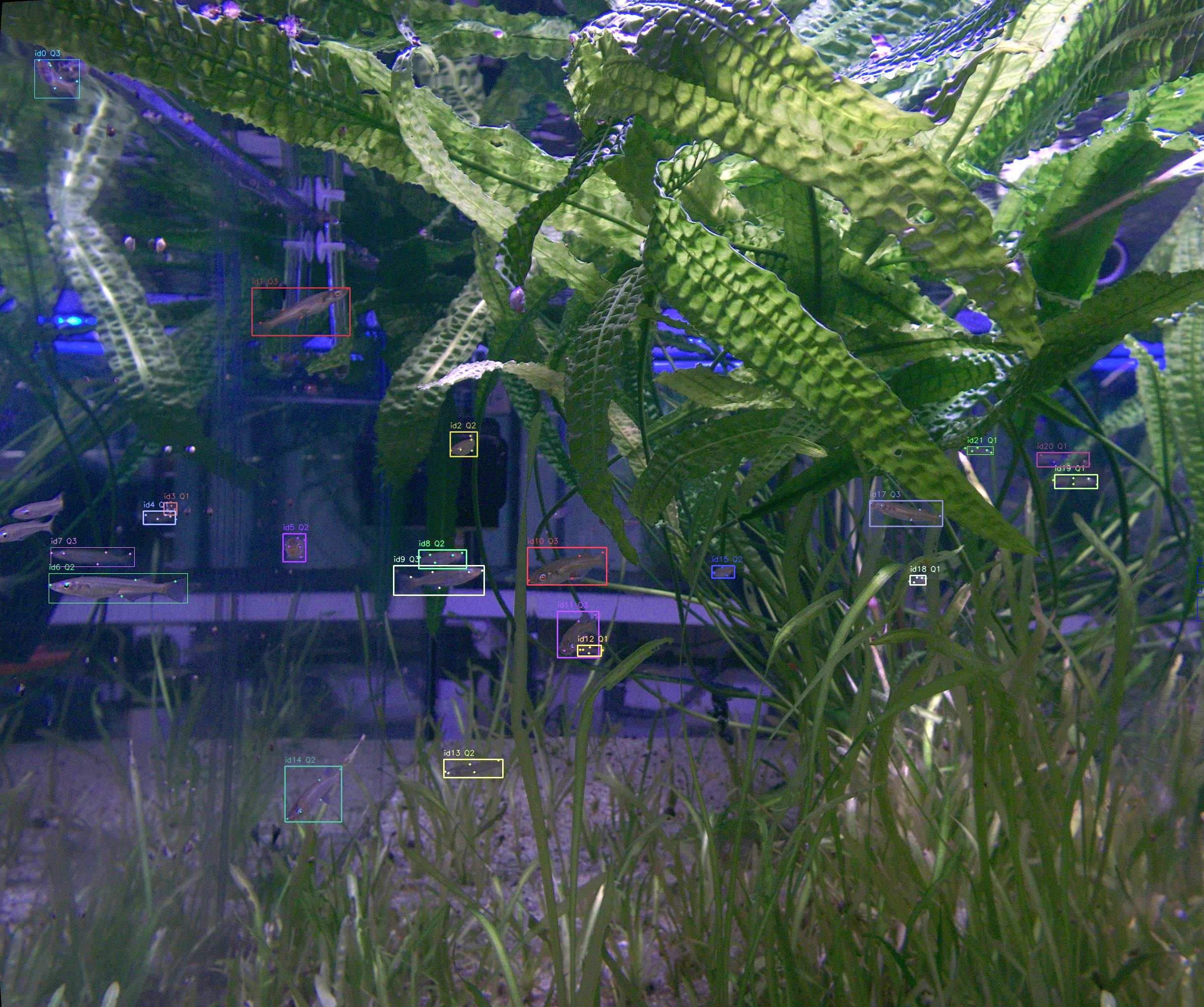} \includegraphics[width=0.475\linewidth]{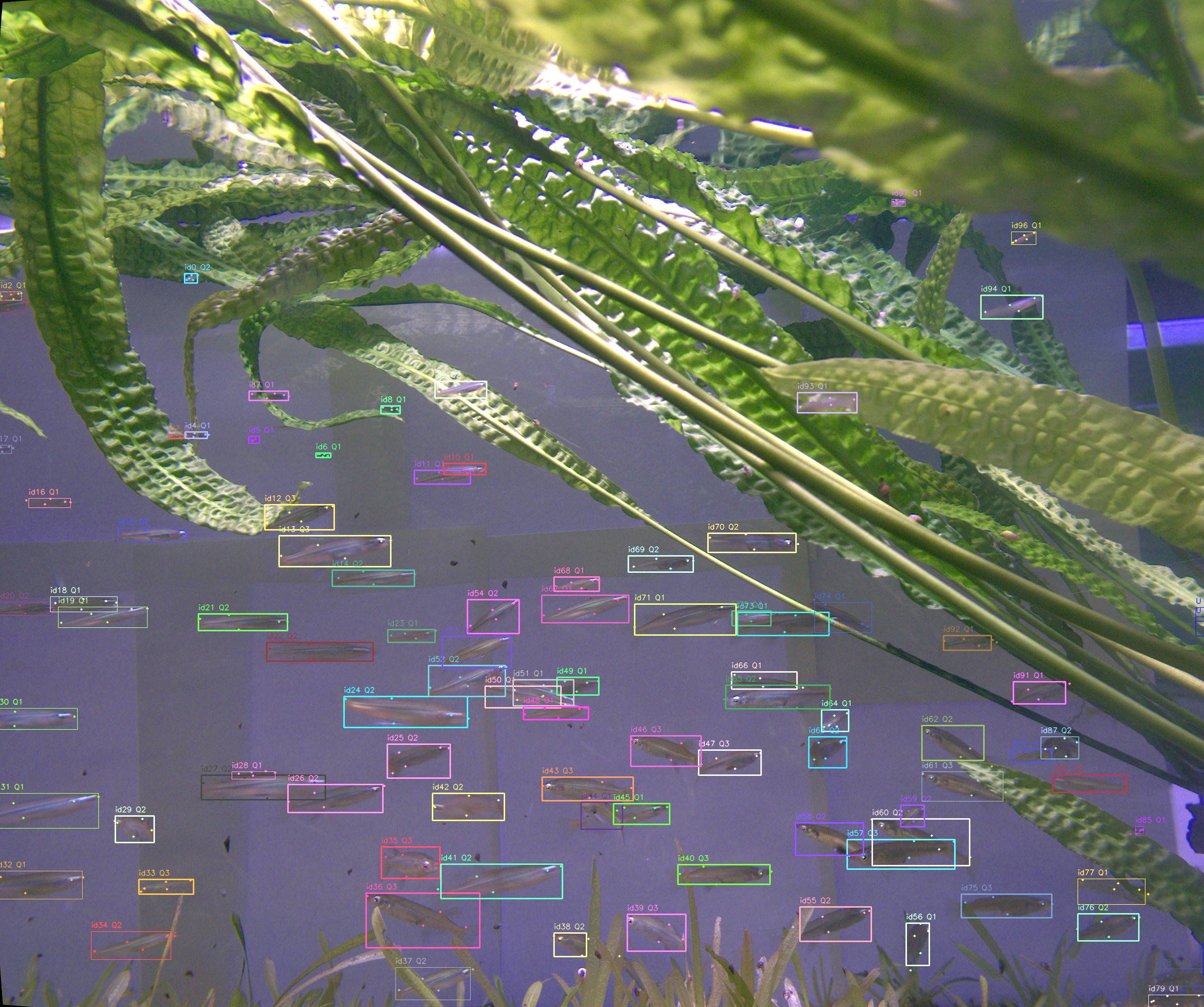}
    \caption{Annotated examples frames of the test-only scenes. In one scene (right), the back of the aquarium was covered with white paper to simulate an aquarium close to a wall.}
    \label{fig:testsetFrames}
\end{figure*}

Our dataset consists of 104 images in total with 4,331 annotated fish instances annotated across several recording sessions. All recordings were captured with the image plane kept approximately parallel to the glass surface. Two sessions, comprising 38 images and 1,903 annotated fish, were designated as test-only data. In one of these sessions, the back wall of the aquarium was covered with white paper to mimic the appearance of home aquariums placed in front of a wall, see Figure~\ref{fig:testsetFrames} for two example frames from the test-only data. 

\begin{figure}
    \centering
    \includegraphics[width=1.0\linewidth]{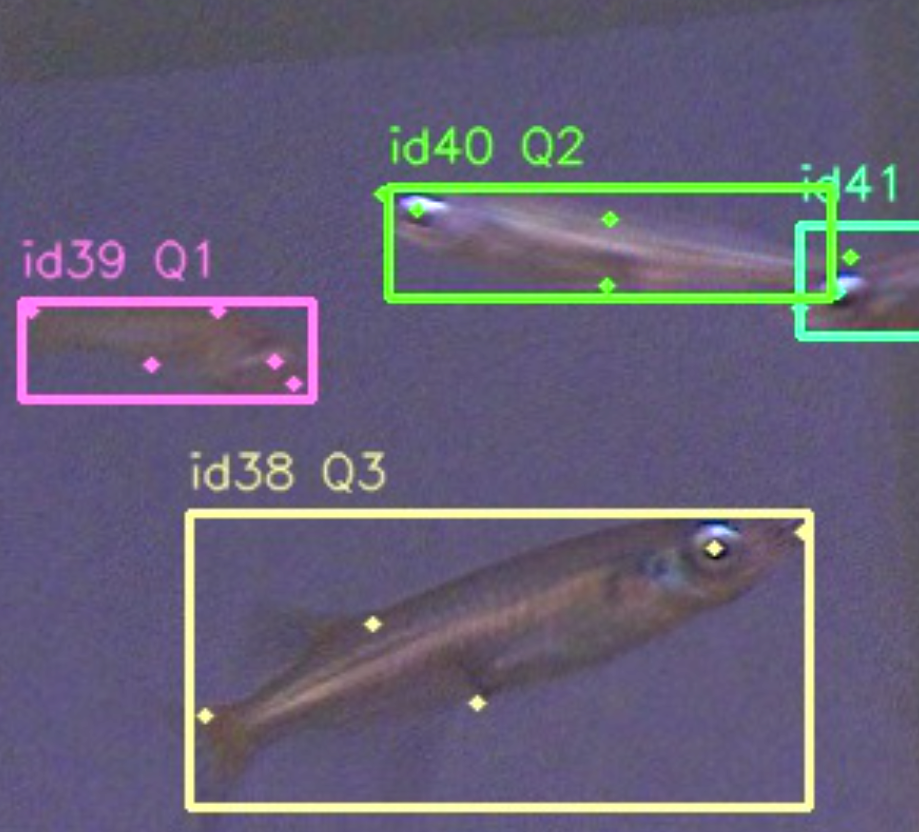}
    \caption{Example of annotated Sulawesi ricefish from our dataset. The annotation includes a bounding box, five keypoints, and a quality class. The fish at the bottom is labeled high quality (Q3) because its keypoints are clearly visible, while the fish at the top (green bounding box) is labeled medium quality (Q2) because its keypoints are hard to recognize due to motion blur. The fish with the pink bounding box is labeled as low quality (Q1) because its keypoints are barely visible due to its low contrast.}
    \label{fig:annotationExample}
\end{figure}
\begin{table}[]
    \centering
    \begin{tabular}{|l|ll|}
         \hline
         &&\\[-1.5ex]
         \# images:&  \multicolumn{2}{l|}{104}  \\
         \# annotated fish: & \multicolumn{2}{l|}{4,331}  \\[1.25ex]
         keypoints per fish: & 5 &(mouth, eye, dorsal fin  \\
                             & &ventral fin, caudal fin)  \\[1.25ex]
         label per fish:  &  &quality of visibility   \\
                            & &(high, medium, low)\\[1.25ex]
        \hline
    \end{tabular}
    \caption{Summary of the Sulawesi Ricefish Stereo Dataset}
    \label{tab:dataset}
\end{table}

Each fish instance is annotated with a bounding box, five anatomical keypoints, and a quality label describing how clearly the fish and its keypoints can be identified. The annotated keypoints correspond to the mouth, eye, dorsal fin, ventral fin, and caudal fin. The quality levels are high, medium, and low: 
\begin{itemize}
    \item High: all five keypoints are clearly visible
    \item Medium: some keypoints are not visible
    \item Low: keypoints are mostly difficult to recognize
\end{itemize}
An example annotation is shown in Figure~\ref{fig:annotationExample}. A short summary of the dataset is provided in Table~\ref{tab:dataset}.

For intrinsic and extrinsic calibration, we captured a checkerboard pattern from multiple viewpoints and processed the image using OpenCV's standard calibration tool for pinhole models and stereo calibration. In both test sessions, the camera's focal point was positioned approximately 40 mm from the aquarium glass. For refraction modeling, we used standard refractive indices of 1.0 (air), 1.5 (glass), and 1.33 (water). 

\section{\uppercase{Methodology}}
\label{sec:methodology}
\begin{figure}[!htb]
    \centering
    \includegraphics[width=0.95\linewidth]{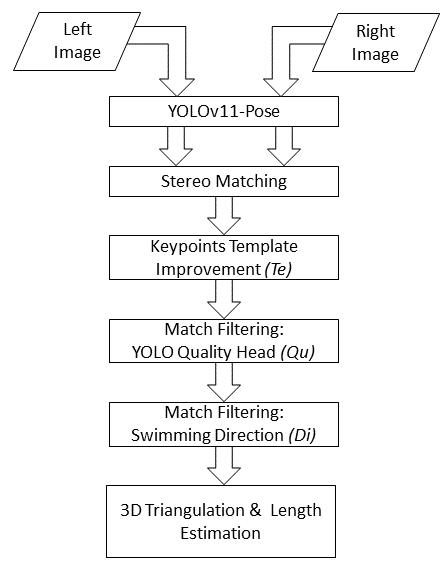}
    \caption{Our approach applies a YOLOv11-Pose separately on the two images of the stereo pair. The detected fish instances are then matched and different components are applied to improve keypoint positions and remove poorly estimated predictions. The individual improvement and quality assessment components are introduces in Section~\ref{subsec:Piepeline}. In brackets are the components' abbreviations used in the results section. }
    \label{fig:approach}
\end{figure}

Our approach consists of several stages that detect and match fish, enforce keypoint consistency, filter unreliable matches and finally compute 3D points and fish lengths, as outlined in Figure~\ref{fig:approach}. In this section, we first describe the training procedure of the YOLOv11-Pose model used to localize fish and keypoints in individual frames. We then introduce the matching, filtering, and 3D measurement pipeline. 

\subsection{YOLOv11-Pose training}
The YOLOv11-Pose architecture predicts bounding boxes and keypoints for detected objects. We extend this model with an additional head to estimate the visibility and overall image quality of each detected fish for reliable length estimation. This additional head is decoupled from the others and consists of two $3\times3$ convolutional filters with Sigmoid Linear Unit activation functions, followed by a  $1 \times 1$ convolutional with three channels, one for each quality level. The head is trained using a cross-entropy loss. 

The training of the YOLO architecture is performed in two stages. In the first stage, the quality estimation head is ignored, while the localization head, the keypoint head, and the backbone are trained. In the second stage, all weights except those of the quality estimation head are frozen and only this head is optimized. This prevents the quality estimator from overfitting and ensures that it learns to assess detection reliability rather than memorizing training samples.

The training was conducted on images with a resolution of $1024 \times 1024$ pixels using an \textit{Nvidia RTX 6000}. We built our method upon the Ultralytics framework~ \citep{yolo11_ultralytics} and used its recommended settings, including the \textit{AdamW}~\citep{LoshchilovH19AdamW} optimizer with a decreasing learning rate after three warm-up epochs, as well as standard augmentation techniques such as horizontal flipping, color-space augmentations, scaling and translations (see \cite{yolo11_ultralytics} for details). We trained and evaluated backbone scales (nano: n, small: s, medium: m, large: l, and extra large: x) of the YOLO architecture. Each training run completed in under 10 minutes, making the approach feasible for small organizations and home users.

\subsection{Length Measurement Pipeline}
\label{subsec:Piepeline}
To measure the length of the fish in a stereo image pair based on predefined keypoints, several processing steps are performed, as described in the following.

First, the YOLOv11-Pose network is applied independently to both images to detect all fish instances and estimate their keypoints along with associated quality levels.

\paragraph{Stereo Matching}
In the next step, stereo matching is performed to associate fish instances across the image pair. For this purpose, we define a cost function with three terms that capture bounding box position as well as keypoint configuration. A greedy assignment strategy is then used to determine the optimal matches.

The first term models the epipolar constraint using epipolar curves. For a fish instance $i$ in one image to map with a fish instance $j$ in the other image, the cost is defined as:
\begin{equation}
    L_p(i,j) = \begin{cases} 
        \frac{\| m_i - epi(m_j; m_i)\|}{150} & \| m_i - epi(m_j; m_i)\| < 150\\
        \inf & \text{otherwise}
    \end{cases}
\end{equation}
where $m_i$ and $m_j$ are the centers of the bounding box in the first and second image in 2D coordinates, respectively, and $epi(m_j; m_i)$ is the closest point on the epipolar curve of $m_i$ to $m_j$.

The second term compares the sizes of the bounding boxes:
\begin{equation}
    L_s(i,j) = \frac{1}{2}\left(\frac{\| w_i - w_j\|}{0.5\cdot (w_i + w_j)} + \frac{\| h_i - h_j\|}{0.5\cdot (h_i + h_j)} \right)
\end{equation}
where $w_i$ and $h_i$ are the width and height, respectively, of the bounding box of fish instance $i$.
The third term models differences in keypoint patterns by calculating the distances between centered keypoints as
\begin{equation}
    L_k(i,j) = \frac{\sum_{k=0}^4 \|(\mathbf{p}_{k;i}-\bar{\mathbf{p}}_{i}) - (\mathbf{p}_{k;j}-\bar{\mathbf{p}}_{j})\|}{{0.5\cdot ( w_i + w_j + h_i + h_j)}}
\end{equation}
where $\mathbf{p}_{k,i}$ is the $k^{th}$ keypoints of the fish instance $i$ in pixel coordinates and $\bar{\mathbf{p}}_i$ is the mean position of all five keypoints defined as
\begin{equation}
\bar{\mathbf{p}}_i = \frac{1}{5}\sum_{k=0}^4 \mathbf{p}_{i;k}.
\end{equation}
Finally, the overall cost function for matching fish $i$ in the first image with fish $j$ in the second image is given by
\begin{equation}
    L(i,j) = \frac{1}{3}\left(L_p(i,j) + L_s(i,j) + L_k(i,j)\right)
\end{equation}

\paragraph{Keypoint Improvement}
After matching fish instances in a stereo image pair, the estimated keypoints are refined using a template matching approach. For each keypoint in the first image, a $21 \times 21$ pixel template is extracted around the keypoint and matched in the second image using the normalized correlation coefficient as similarity metric. The search area is constrained to $\pm 30$ pixels around the estimated keypoint position in the second image and must be within a distance of 5 pixels from the corresponding epipolar curve.

\paragraph{Match Filtering}
Finally, matches are filtered to remove inaccurate estimates and retain only reliable and accurate keypoint correspondences. First,  fish instances with low or medium quality predictions are discarded. Two additional filters account for the fish's orientation, as fish swimming directly towards or away from the camera make accurate keypoint localization difficult. Fish with a bounding box aspect ratio, defined as width divided by height, of less than 1.5 are removed. Additionally, the swimming direction relative to the camera's optical axis is estimated by the angle between a line connecting the fish's mouth and tail fin and the optical axis. Fish with an angle of less than $45^\circ$, i.e., swimming roughly towards or away from the camera, are excluded from the set of valid matches.

\section{\uppercase{Results}}
\label{sec:results}
We evaluate our approach on our Sulawesi Ricefish Stereo Dataset described in Section~\ref{sec:dataset}. The YOLOv11-Pose network was applied to images at a resolution of $1024\times1024$, consistent with the training setup, while subsequent processing was performed on images in their original size.

Our evaluation addresses three aspects of the proposed approach. First, we assess the performance of the YOLOv11-Pose's quality level prediction, as it is an essential component for identifying reliable matches. Second, we evaluate the stereo matching performance, since correct matches are required for accurate fish size estimation. Finally, we examine the accuracy of the complete fish size estimation pipeline. We define fish length as the distance from the mouth to the caudal fin. The accuracy is evaluated reporting root mean squared error (RMSE) of the estimated fish length. 

We further perform an ablation study to examine the effect of different components of our pipeline. In particular, we assess the impact of the visibility quality filter, the keypoint refinement via template matching, and swimming direction check on the overall system performance. 

\begin{table}[]
    \centering
    \begin{tabular}{c|ccc}
ground-truth     & \multicolumn{3}{c}{predicted quality}\\
quality     & low & medium & high\\
    \hline
       & \multicolumn{3}{c}{backbone: \textbf{nano}}\\
low    &0.870   &  0.497   &  0.161\\
medium &0.129   &  0.440   &  0.435\\
high   &0       &  0.063   & 0.403 \\[1.5ex]

       & \multicolumn{3}{c}{backbone: \textbf{small}}\\
low    &0.851    & 0.541    & 0.141\\
medium &0.141    & 0.402    & 0.459\\
high   &0.007    & 0.057    & 0.400\\[1.5ex]

       & \multicolumn{3}{c}{backbone: \textbf{medium}}\\
low    & 0.878   &  0.510  &  0.117\\
medium & 0.108   &  0.443  &  0.458\\
high   & 0.014   &  0.047  &  0.425\\[1.5ex]

       & \multicolumn{3}{c}{backbone: \textbf{large}}\\
low    & 0.872   & 0.559  &  0.139\\
medium & 0.120   & 0.410  &  0.451\\
high   & 0.008   & 0.031  &  0.410\\[1.5ex]

       & \multicolumn{3}{c}{backbone: \textbf{extra large}}\\
low    & 0.870 &  0.516   &  0.142\\
medium & 0.131 &  0.420   &  0.438\\
high   & 0     &  0.065   &  0.420\\[1.5ex]
             \end{tabular}
    \caption{Confusion matrix for quality level prediction of different YOLO backbones}
    \label{tab:confusionMatrix}
\end{table}

\paragraph{Quality Level Prediction}
Table~\ref{tab:confusionMatrix} shows the normalized confusion matrix for the quality-level prediction for different YOLO backbone scales. The choice of backbone scale has a marginal effect on the performance, and the overall trends are consistent for all YOLO configurations. As we are primarily interested in fish with clearly recognizable keypoints, the performance of the quality class \textit{high} is the most relevant factor. In particular, the number of false positive (instances predicted as high quality but labeled as medium or even low) is important. Missing a fish with visible keypoints can be compensated by considering additional instances, for example from adjacent frames, but including an instance with poorly estimated keypoints can distort the length distribution. 

Across all configurations, only a small number of fish labeled as high quality are missed, indicating that the false-rejection rate for truly high-quality samples is negligible. However, only about 40-42.5\% of the detections predicted as high quality are actually annotated as such. Most of the remaining high-quality predictions  fall into the medium-quality category, which is not critical given that the boundary between the classes is not objectively measurable and several instances could reasonably also be annotated as high quality. Only about 15\% of the fish predicted as high quality are labeled as low quality. 

\paragraph{Stereo Matching Accuracy}
\begin{table}[!htb]
    \centering
    \begin{tabular}{c@{\hspace{2pt}}|c@{\hspace{2pt}}|c@{\hspace{2pt}}|ccccc}
    \multicolumn{3}{c|}{Used}& \multicolumn{5}{c}{}  \\
    \multicolumn{3}{c|}{Com-}& \multicolumn{5}{c}{Yolo-Backbone}  \\
    \multicolumn{3}{c|}{ponents}& \multicolumn{5}{c}{}  \\
    \textit{Qu} & \textit{Te} & \textit{Di} &  n    &   s    &   m    &   l    &   x \\
    \hline
     &  &  & 11.6 &  17.7 &  19.8 &  13.9 &  16.2 \\
    X &  &  & 4.6 &  1.5 &  2.4 &  2.8 &  \textbf{1.1} \\
     & X &  & 5.4 &  10.7 &  11.7 &  7.3 &  9.2 \\
    X & X &  & 3.4 &  1.6 &  1.9 &  2.5 &  1.3 \\
     &  & X & 6.6 &  2.6 &  3.9 &  1.3 &  3.2 \\ 
    X &  & X & 6.1 &  1.7 &  \textbf{1.1} &  1.9 &  2.0 \\  
     & X & X & 4.3 &  2.1 &  2.8 &  3.3 &  2.8 \\
    X & X & X & 3.7 &  1.8 &  2.2 &  2.9 &  2.2 \\[1.25ex]    
    \multicolumn{3}{c|}{Average} & 5.7 &  5.0 &  5.7 &  4.5 &  4.7
    \end{tabular}
    \caption{Percentage of Bad Matches [\%]. Best performing settings are marked in bold. The components’ abbreviations \textit{Qu}, \textit{Te}, and \textit{Di} refer to Quality Head, Template Matching, and Swimming Direction, respectively. See also Figure~\ref{fig:approach} and Section~\ref{subsec:Piepeline}.}
    \label{tab:AccuracyStereo}
\end{table}
Table~\ref{tab:AccuracyStereo} shows the number of correctly matched stereo instances. To this end, we calculated the percentage of bad matches as the amount of bounding box pair that cannot be mapped to a pair in the ground truth data. The component \textit{Qu} (for quality filter) denotes the selection of filtering of samples using the YOLOv11 quality estimation head to remove low and medium quality instances, \textit{Te} (for template matching) refers to keypoint refinement via template matching along epipolar curves, and \textit{Di} (for direction filter) corresponds to filtering instances based on the bounding box ratio and the orientation of the fish with respect to the camera. 

The best performance is achieved with the m-scale backbone and the components \textit{Qu} and \textit{Di} and the x-scale backbone with only the \textit{Qu} component. In general, the percentage of incorrect matches is low, when either the \textit{Qu} and \textit{Di} is applied to remove poorly estimated instances.

\paragraph{Accuracy of the Length Estimation}
\begin{table*}[!htb]
    \centering
    \begin{tabular}{c|c|c|ccccc|ccccc}
    \multicolumn{3}{c|}{Used} & \multicolumn{5}{c}{} & \multicolumn{5}{c}{} \\
    \multicolumn{3}{c|}{Components} & \multicolumn{5}{c}{Scene noBG} & \multicolumn{5}{c}{Scene wBG} \\
    \textit{Qu} & \textit{Te} & \textit{Di} &  n    &   s    &   m    &   l    &   x    &  n    &   s    &   m    &   l    &   x    \\
    \hline
     &  &  & 22.62 &  18.97 &  17.78 &  16.08 &  18.57 &  20.14 &  18.32 &  18.92 &  19.57 &  18.19 \\
    X &  &  & 17.87 &  18.75 &  14.11 &  12.11 &  14.16 &  18.78 &  16.84 &  15.86 &  16.38 &  15.19   \\
     & X &  & 20.39 &  16.82 &  21.21 &  17.51 &  20.06 &  19.29 &  17.68 &  18.33 &  17.59 &  17.95 \\
    X & X &  & 17.85 &  15.09 &  20.79 &  16.36 &  19.56 &  16.22 &  15.54 &  11.78 &  11.75 &  12.64 \\
     &  & X & 14.17 &  15.43 &  11.64 &  11.04 &  11.22 &  17.33 &  15.59 &  15.41 &  16.32 &  14.27  \\
    X &  & X & 12.82 &  15.39 &  \textbf{6.11} &  9.71 &  12.61 &  15.75 &  12.97 &  13.12 &  15.61 &  11.87  \\
     & X & X & 15.74 &  9.73 &  11.10 &  11.79 &  11.82 &  15.12 &  13.44 &  12.07 &  13.02 &  15.21  \\
    X & X & X & 12.20 &  7.25 &  11.17 &  9.45 &  8.46 &  13.62 &  11.21 &  \textbf{9.22} &  10.59 &  9.51  \\
    \end{tabular}
    \caption{Root Mean Squared Error of length estimation for different configurations in millimeters. Best numbers per scene are marked in bold. The components’ abbreviations \textit{Qu}, \textit{Te}, and \textit{Di} refer to Quality Head, Template Matching, and Swimming Direction, respectively. See also Figure~\ref{fig:approach} and Section~\ref{subsec:Piepeline}.}
    \label{tab:Accuracy}
\end{table*}
The overall accuracy, reported as root mean squared error (RMSE) of the estimated fish length, is given in Table~\ref{tab:Accuracy} and evaluated individually for the two scenes from the test-only set. We refer to the scene with the white background as \textit{scene wBG} and to the other as  \textit{scene noBG}. A sample image from \textit{scene noBG} and \textit{scene wBG} is shown in Figure~\ref{fig:testsetFrames} left and right, respectively. For the abbreviations of the used component in the table, see the previous paragraph.

The quality-based filtering improves the RMSE in nearly all cases, with a reduction of nearly 50\%  in some cases, e.g., for Scene \textit{scene noBG} with no template improvement and an active direction-based filter. The direction-based filtering also improves accuracy in nearly all cases, indicating that this component primarily removes problematic samples. The effect of the template-matching for the keypoint improvement strongly differs between the two scenes. For scene \textit{scene noBG}, the refinement often impairs the accuracy, whereas for scene \textit{scene wBG} it always leads to a significant improvement. 

This behavior can be explained by the different backgrounds of the scenes. It directly affects the template matching, because background elements are partly included in the template and the matching method attempts to align these patterns as well. Since the background elements are more visually dominant in Scene \textit{scene noBG}, the refinement is more easily misled.

Another noteworthy factor lies in the nature of the Sulawesi ricefish. They prefer sheltered areas near walls, plants, or other aquarium decoration. In Scene \textit{scene wBG}, the fish tend to remain close to the back of the aquarium, similar to how fish behave in home aquariums placed in front of a solid wall. In contrast, in Scene \textit{scene noBG} more often hide among plants or near other objects and thus increase the complexity of the template matching.

The size of the YOLOv11 backbone has the same effect for both scenes. The best results are obtained with the \textit{m}-scale backbone. Smaller backbones appear to lack sufficient capacity for accurate keypoint estimation, while larger one may suffer from overfitting. 

Because this metric does not reflect how many predicted instances could not have been matched to a ground truth annotation, we also studied how many predictions could now have been matched to annotated data. The matching was performed using a greedy strategy based on the distance between bounding box centers, with a maximum allowed distance of 30 pixels. Without the quality filter (\textit{Qu}) and without the orientation (\textit{Di}) filter, several predictions could not be matched to any ground truth annotations. When these filters were enabled, however every prediction could be matched to a corresponding ground truth sample.

\paragraph{Runtime Performance Evaluation}
We evaluated our length measurement system on a computer equipped with an AMD EPYC 9654 processor and an Nvidia GeForce RTX 6000 GPU. In this setting, our approach processes about five frame pairs per second. The actual runtime for a single image pair varies strongly with the number of detected fish instances. The computational bottleneck is the keypoint improvement via template matching, which accounts for approximately 75\% of the total runtime. The most time-consuming operation is the calculation of the distances between the possible template shifts and the closest points of the corresponding epipolar curve as all line segments are considered. 

\section{\uppercase{Conclusions}}
\label{sec:conclusion}
Measuring the growth of fish can be a helpful tool to monitor their health and to detect diseases, improper nutrition, and other harmful environmental influences in fish cultivations. In this paper, we proposed a system that robustly measures the size of fish in a home aquarium using a stereo setup, addressing the challenges of refractive distortions, inconsistent keypoint predictions, and unreliable stereo correspondences through learned quality assessment and refraction-aware triangulation.

Our method uses a YOLOv11-Pose network to predict bounding boxes and predefined keypoints of fish based on single images. In further steps, the fish instances of a stereo image pair are fused, filtered to discard poorly estimated keypoints and finally, the keypoints are used to a 3D estimation of predefined keypoints on the fish. To consider the image quality of fish in aquariums, which can strongly vary due to lack of light, fast motion, or occlusion by other fish, plants, and aquarium decoration, we extended the YOLOv11-Pose by an additional head to predict the quality. This predicted quality is combined with further filters that analyze the orientation of the fish instances. A fish heading towards to or away from the camera is rejected because the keypoints are in these cases usually hard to detect. In addition, a template matching restricted by an epipolar constant was included in the system.

We evaluated our system on a challenging dataset of Sulawesi ricefish that grow only up to 80 mm. Our proposed filter methods have been proved to successfully remove poorly estimated fish instances and thus significantly reduce error rates caused by poorly estimated keypoints and wrong stereo matches. Whether template matching improves or impairs the accuracy of our systems strongly depends on the background. If the aquarium is placed in front of a wall it can significantly improve accuracy, but otherwise tend to impair it. 

As long exposure times are often required to capture well-illuminated images of aquariums, motion blur is a prevalent artifact that impairs accurate keypoint estimation. 
In future work, we plan to integrate a model-based deblurring approach \citep{Seibold2017MotionBlurTracking} to mitigate motion blur and improve the robustness and accuracy of our keypoint estimation. Furthermore, we plan to improve the runtime performance of the keypoint improvement via template matching by replacing the time-consuming computation of distances between image points and epipolar curves with an efficient approximation.

\section*{ACKNOWLEDGEMENTS}
This work was partly funded by the German Federal Ministry for Economic Affairs and Energy  (FischFitPro, grant no. 16KN095536) and the Federal Ministry of Research, Technology and Space (REFRAME, grant no. 01IS24073A).

\bibliographystyle{apalike}
{\small
\bibliography{example}}

\end{document}